# Digital Quadruplets for Cyber-Physical-Social Systems based Parallel Driving: From Concept to Applications

Teng Liu, Xing Yang, Hong Wang, Xiaolin Tang, Long Chen, Huilong Yu, Fei-Yue Wang

*Abstract—* **Digital quadruplets aiming to improve road safety, traffic efficiency and driving cooperation for future connected automated vehicles are proposed with the enlightenment of ACP based parallel driving. The ACP method denotes Artificial societies, Computational experiments and Parallel execution modules for cyber-physical-social systems. Four agents are designed in the framework of digital quadruplets: descriptive vehicles, predictive vehicles, prescriptive vehicles and real vehicles. The three virtual vehicles (descriptive, predictive, and prescriptive) dynamically interact with the real one in order to enhance the safety and performance of the real vehicle. The details of the three virtual vehicles in the digital quadruplets are described. Then, the interactions between the virtual and real vehicles are presented. The experimental results of the digital quadruplets demonstrate the effectiveness of the proposed framework.**

## I. Introduction

Autonomous vehicles have been extensively researched and developed in the past few decades to address the increasing demands on safer driving and more efficient and intelligent transportation systems in the past years [1]. Autonomous vehicles would be expected to significantly improve road safety. From autonomous vehicles in the DARPA challenge to the self-driving cars developed by Google, Uber and other companies [2], substantial technologies have been advanced covering from level one towards level four+ automation. However, considerable challenges in perception, decision, planning and control still need to be tackled before large scale implementation of autonomous vehicles, partially due to significant complexity and uncertainties of driving environment. This also calls for novel frameworks in connected autonomous driving. Motivated by this, the parallel driving, a cloud-based cyber-physical-social systems (CPSS) framework has been developed and implemented in multiple real-world driving scenarios [3,11].

In this article, we advance the paralllel driving concept by formulating the whole system into four different interactive parts, named as digital quadruplets. Within digital quadruplets each of the connected real vehicles is accompanied by three virtual vehicles or 'guardian angels'. These three virtual vehicles are developed in the artificial world and they are named as: descriptive vehicle, predictive vehicle, and prescriptive vehicle. The descriptive vehicle focuses on modeling dynamic behaviors of real vehicles. The predictive vehicle aims to forecast the corresponding consequences generated by the current control actions in a predefined time span. The objective of the prescriptive vehicle is to guide the real vehicle regarding what actions should be taken in incoming driving scenarios.

To introduce the digital quadruplet for parallel driving clearly, the remaining paper is arranged as follows. Section II describes the framework of parallel driving and the structure of digital quadruplets. The three virtual 'guardian angels' in the artificial world and the evaluation of the parallel driving system are introduced in Section III. In Section IV, the real-time interaction behaviors between the virtual and real vehicles are illustrated, with the experimental validation of the parallel driving system being performed and discussed, followed by the conclusion remarks in Section V.

## II. Digital Quadruplets in Parallel Driving

This section defines the digital quadruplets in cyber-physical-social system based parallel driving. First, the concepts of CPSS and ACP method (Artificial societies, Computational experiments and Parallel execution method) are concisely introduced. Based on these, the framework of parallel driving is given. Finally, the digital quadruplets in the parallel driving are discussed.

### A. CPSS and ACP Method

Human involvements and social characteristics impose an additional dimension to traditional cyber-physical systems (CPS), which leads to the framework of CPSS [1-3]. Rapid

T. Liu, and X. Tang are with State Key Laboratory of Mechanical Transmissions, College of Automotive Engineering, Chongqing University, Chongqing, 400044, China. (email: tengliu17@gmail.com, tangxl0923@cqu.edu.cn)
Y. Xing is with the Department of Mechanical and Aerospace Engineering at Nanyang Technological University, Singapore (email: yxing_edu@163.com)
H. Wang is with School of Vehicle and Mobility, Tsinghua University, Beijing, China. (email: hong_wang@tsinghua.edu.cn, Corresponding Author)
L. Chen is with Waytous Inc., Qingdao Shandong 266109, China. (email: chenl46@mail.sysu. edu.cn).
H. Yu is with the Department of Mechanical and Mechatronics Engineering, University of Waterloo, N2L 3G1, Canada. (email: huilong.yu@uwaterloo.ca)
F.-Y. Wang is State Key Laboratory for Management and Control of Complex System, Institute of Automation, Beijing, China (e-mail: feiyue@ieee.org).

2development in artificial intelligence, big data, and communication technologies enables development of CPSS to attract more and more attention in recent years. The integration of critical elements in the CPSS contain the physical, mental and artificial worlds. It indicates that the ACP method can optimize and manage the complex systems in CPSS via exploiting knowledge in the artificial world to guide the mental and physical worlds.

The ACP method means the artificial societies (A), computational experiments (C) and parallel execution (P), which was proposed by Fei-Yue Wang since 2004 [3-6]. To describe concisely, A is usually used for modeling the complex systems, C is applied to calculate and analyze the responses and P is utilized for control and management. These three components of ACP can also be mapped into three parallel worlds which are called physical, mental and artificial worlds. In a parallel system, ACP can be activated in both the artificial system and the real system to handle complex control problems [7, 8].

*B. Parallel Driving*

Autonomous vehicles are experiencing a rapid development [9, 10]. However, an individual intelligent vehicle is usually not able to recognize social behaviors related to driving on public roads. It is a big challenge for the vehicle to understand the meaning of other vehicles' behaviors or intentions and to figure out how to safely and efficiently interact with them. Hence, CPSS-based parallel driving was proposed to achieve improved safety, performance and efficiency (Fig. 1) [11].

Different from the conventioanl connected automated driving, parallel driving includes artificial drivers and artificial vehicles. The parameters and information of the real and artificial vehicles co-exist within the three parallel domains [12]. Specifically, the real driving exists in the physical world, which consists of the physical behavior of the real vehicle and real driver. The cognitive behaviors of the real and artificial drivers located in the mental world, including driver attention, intention and attributes. There are two layers in the artificial world, the first layer is the artificial driver and artificial vehicle, and the second layer is the information of people, location and technologies [13-15].

The modeling of the vehicle, driver, environment and so on will be built in the artificial world. The artificial vehicles can be trained in both normal and corner driving scenarios, and the generated control commands could guide the real vehicles in the physical world. Furthermore, the data from the real vehicle will be uploaded to improve the accuracy of the artificial modeling. These processes can be realized by the special computational experiment and parallel execution methods [16-17]. Different types of the virtual vehicles can co-exist in the artificial world to communicate with the real vehicle, which are discussed in the next section.

*C. Digital Quadruplets in Parallel Driving*

Parallel driving framework is enhanced to include four interactive parts and they are real vehicles and three virtual vehicles, named as three guardian angles A, B and C. The three guardian angels are named as guardian angel A: descriptive, guardian angel B: predictive, and guardian angel C: prescriptive vehicles. These vehicles can be regarded as the 'guardian angels' of the real vehicles in parallel driving. Combining with the real autonomous vehicle in parallel driving, they are called the digital quadruplets of parallel driving. As depicted in Fig. 2, different real vehicles in the physical world could be with any automation level, from Level 0 to Level 5. They interact with each other to achieve safe and efficient management [18] (e.g., fleet management [19]).

Based on the historical driving data and current states, the first guardian angel, descriptive vehicle aims to describe the operation of real vehicles [20] accurately. For example, how to model the autonomous vehicle powertrain, how to establish various driving scenarios for real vehicles, how to build the environment modeling and what data should be selected for communications between the physical and artificial worlds. By solving these challenges, this virtual vehicle can form as a self-consistent system to represent the real driving and real vehicle in the artificial world.

The second virtual guardian angel, the predictive vehicle can be treated as multi-agent learning systems, which use diverse computation experiment approaches (e.g., machine learning [21] and software analytics [22]) to realize self-calibration and self-sublimating. The first step is self-calibrating, based on the modeling, data and knowledge from the descriptive vehicles and real vehicles, the predictive vehicles can improve the generative virtual models and evaluate the real physical models. For example, to perform the same controls into the descriptive vehicles and real vehicles and make them to output the similar states. The second step is self-sublimating, using the artificial intelligence (AI) technologies to make all the virtual vehicles traverse different testing scenarios, for example, emergency scenarios and extreme weather conditions, and then to boost themselves and to guide the real vehicles.

Finally, the third guardian angel, the prescriptive vehicle decides two aspects of concerns, the first one is whether the controls learned from the predictive vehicles can be used into the physical world, and the second one is how the virtual vehicles interact with the real ones. To address the first problem, the safety-aware, behavior-aware and situation-aware-based learning methods (e.g., apprenticeship learning [23]) can be leveraged to evaluate the controls, policies and strategies from predictive vehicles. To solve the second problem, planning and control, as well as human-machine interface techniques will be employed to link the virtual and real vehicles. In the following Sections, we present the details of the four digital vehicles in parallel driving.

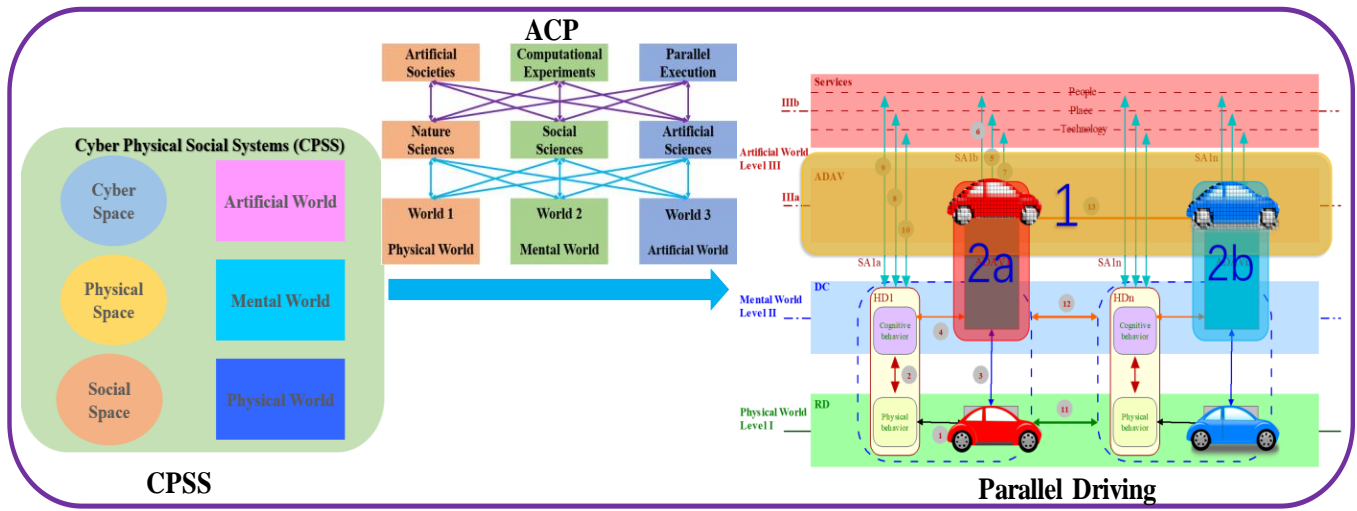

Fig. 1. CPSS-based parallel driving architecture.

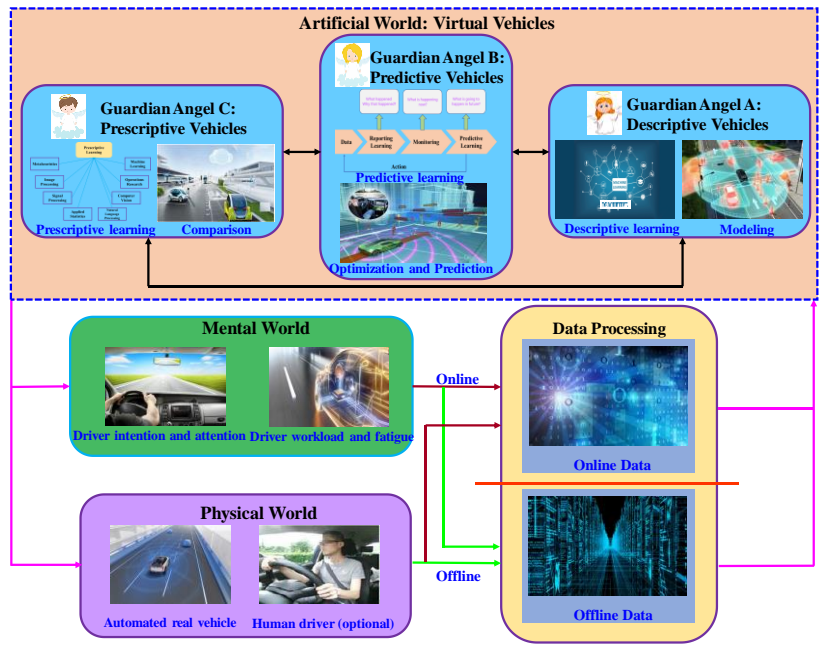

Fig. 2. The proposed framework of the digital quadruplets for parallel driving.

### III. FOUR VEHICLES IN DIGITAL QUADRUPLETS

#### A. Guardian Angel A: Descriptive Vehicle

The first component of the digital quadruplets is called the descriptive vehicle. Its construction relies on the descriptive learning theory. The nature behind the descriptive learning is to make strategies that can describe how the learning is generated and how to learn a model that can be used to mimic and describe the real-world vehicle. The descriptive vehicle is one of the core components in parallel driving which is responsible for learning of real-world vehicle dynamics based on limited prior knowledge and observations.

The descriptive vehicle in the artificial world needs to learn the pattern of the dynamics and behaviors of the real vehicle. The descriptive vehicle is parallel to the real vehicle and will run asynchronously. The descriptive vehicle is an active counterpart of the real vehicle, which provides a playground to explore the space of the state and potential response of the real agent in a more efficient and less harmful manner. The descriptive vehicle could be built based on hybrid model-based approach and descriptive learning theory.

#### B. Guardian Angel B: Predictive Vehicle

To be an intelligent vehicle of high automation level, the vehicle needs to perceive the states of the world first to make accurate predictions and planning. For example, predicting the consequences of a sequence of control commands provided by the onboard controller. This definition of predictive learning is also meant by unsupervised learning and has been researched for many years.

The predictive vehicles can achieve different purposes in prediction and planning, and one of the algorithms in the predictive vehicles is utilized to predict different variables:

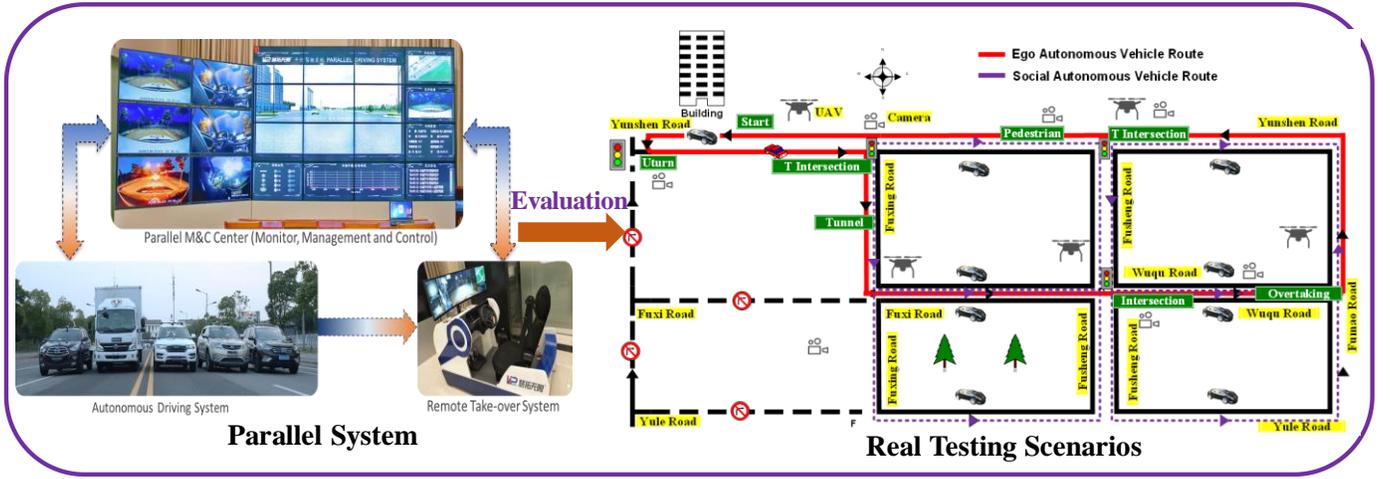

Fig. 3. Parallel driving system tested in the real driving scenarios.

the fuzzy encoding module (FEM) [24, 25]. This virtual vehicle predicted by FEM can predict many states of the real vehicles based on the historical data, such as vehicle speed, power demand, engine speed, motor current and so on. By doing so, the vehicles can adjust the control actions according to the prediction and thus run more safely and efficiently [26].

### C. Guardian Angel C: Prescriptive vehicle

The prescriptive vehicle is responsible for guiding the real vehicle to make actions and achieve the target outcomes, which aims to generate a series of proper action strategies to help the real vehicle deal with the complex traffic situation. The prescriptive vehicle is concerned about whether the policy learning in the parallel system can be properly utilized in the real world. By and large, the prescriptive vehicle can be treated as a special kind of predictive vehicle. However, the prescriptive vehicle must be smarter than the predictive vehicle since the prescriptive vehicle has to learn the long (or short) term dependency between the taken actions and the response in case to guide the real vehicle when facing complex situations. As the predictive vehicle enables the predictions of what will happen in the future, the prescriptive vehicle can take the predictions from multiple parallel predictive vehicles to evaluate which action contributes to the maximum rewards and generate the target consequence [27].

### D. Evaluation of Parallel Driving System

The real vehicles in parallel driving can be at different automation levels, either with human driver in the loop or not. The number of them is variable. This vehicle can be an individual one in the parallel driving system and interacts with the virtual vehicle in the artificial world. Furthermore, they also can constitute a vehicle fleet and communicate with each other through vehicle-to-vehicle (V2V) communications.

Figure 3 shows the digital quadruplets were tested and evaluated in the real driving scenarios. The parallel driving system includes the several real autonomous vehicles and three virtual guardian angels in the parallel M&C (Monitoring, Management and Control) center. The modeling for human driver, perception and vehicle powertrain were conducted in the descriptive vehicle. In the predictive vehicle, we evaluated the performance of real vehicles in different scenarios, such as crossing intersection and tunnel, overtaking, U-turn and yielding right of way for pedestrians. In the prescriptive vehicle, the real vehicle was controlled within the parallel driving architecture, and if needed, taken over by the remote take-over system upon system request. In the next section, we use the collected data to describe the different vehicles in the digital quadruplets and how they work together to improve the safety and efficiency of autonomous driving.

## IV. EXPERIMENT RESULTS IN PARALLEL DRIVING

The parallel driving system was developed and evaluated in Changshu, Jiangsu province, China. Based on the experimental data, this section illuminates that the virtual vehicles are co-operated with the real vehicles to achieve more efficient and safer driving. The simultaneous interactions between the real and virtual vehicles can follow the simulation protocol of the digital quadruplet that we defined in the last section very well. Specifically, the real-time interaction flow is illustrated in Fig. 4, where the four kinds of vehicles interact and cooperate with each other dynamically. The naturalistic driving data can be transferred to each of the digital vehicles to provide real-world guidance and to refine the design and learning process of digital vehicles. Combined offline and online trainings cold be employed for three virtual vehicles. Meanwhile, information and knowledge share are allowed among digital vehicles, the knowledge from the virtual vehicles will be sent to the real vehicle to optimize the real-world driving strategies [28].

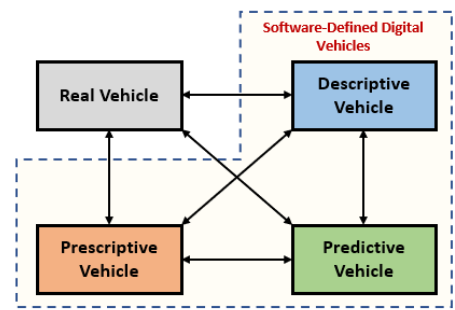

Fig. 4. Dynamic simulation protocol for digital quadruplets vehicles.

The simulation validation of the digital quadruplets-based parallel driving can be described as follows. First, the descriptive vehicle will be designed in the simulation environment considering the key simulation parameters of the real vehicle, such as the kinematic, motion dynamics, driving styles, and traffic context, etc. The traffic environment information was recorded by Lidar and radar, and it is utilized to build artificial scenarios (the software used here is PanoSim), as shown in Fig. 5. The precise modeling of the descriptive vehicle is the cornerstone of the following simulation studies, as the descriptive vehicle is responsible for behavioral learning of the real-world vehicle based on the received driving data and observation of social behaviors. As mentioned in Fig. 3, several real vehicles were driven by human drivers, and the three virtual vehicles were learned from the real driving in the artificial world.

Second, the predictive vehicle of the quadruplets can be developed in the virtual world to simulate challenge scenarios and generate optimized driving strategies. Figure 6 shows the pictures taken by the real cameras and virtual ones in the physical and artificial worlds, respectively. Different driving scenarios (e.g., emergency situations) can be imitated in the artificial world to improve the safety and efficiency of the real vehicles with the guardian angels. Extra vehicles and pedestrians can be added to train the decision-making and planning modules in the artificial world. The computation experiments for the predictive vehicle is not governed by the limited naturalistic driving data, both common scenarios and uncommon critical scenarios can be simulated so that much more knowledge and strategies can be obtained.

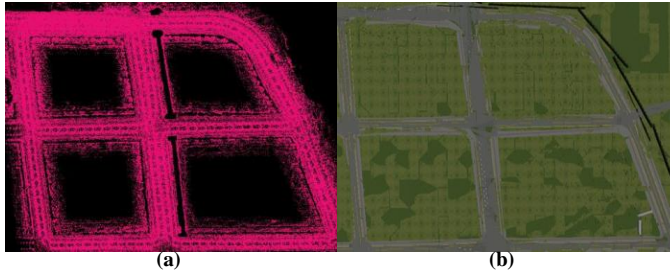

Fig. 5. Road information in the real driving experiment (a) Roadmap generated by the onboard Lidar information; (b) Roadmap built in the software-defined artificial world.

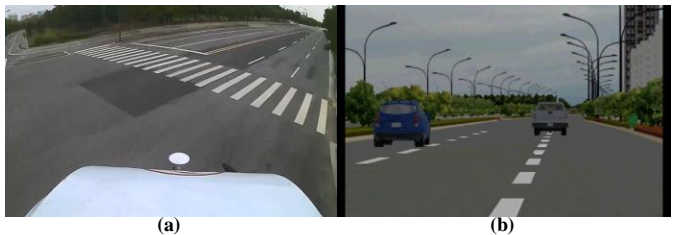

Fig. 6. First perspective of the real and virtual vehicles (a) Picture taken by the real onboard camera; (b) First perspective of the virtual vehicle in Panosim.

Last, the prescriptive vehicle was constructed based on the descriptive and predictive vehicles to guide the driving and to increase the intelligence of the real vehicle. For example, the real controller can generate commands to control the real vehicles, and predictive vehicles in the artificial world can also guide the virtual vehicles in various scenarios. Figure 7 indicates the velocity and steering angle in real and virtual vehicles, wherein the big steering angles mean sharp right turning on the roads. The data from the real vehicle is represented by the blue line, and the data from the virtual vehicles are represented by the red line. It is obvious that the speed trajectories and time intervals are different in these two cases, which are affected by the driving scenarios. Several emergency situations are added into the virtual scenarios, and thus the speeds of virtual vehicles sometimes equal to zero to avoid collisions. Considering the waiting time in the virtual vehicles, the time interval is longer than that of the real vehicles. The prescriptive vehicle will optimize the driving strategies under these emergency situations and provide a series of action strategies to help the real vehicle deal with unseen complex traffic situations.

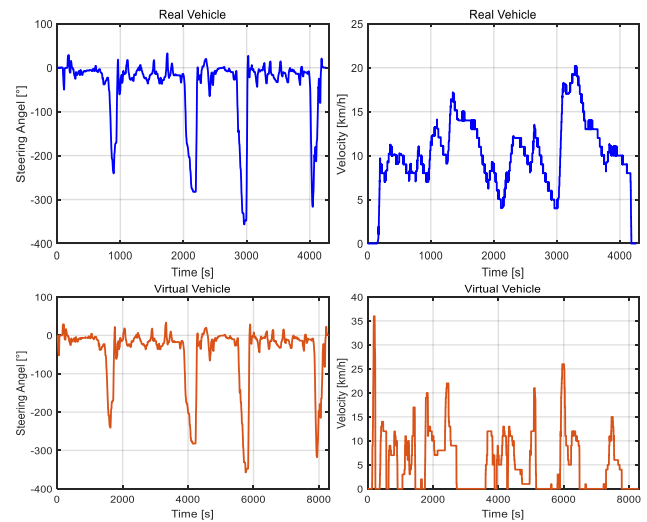

Fig. 7. Velocity and steering angel commands in real and virtual vehicles

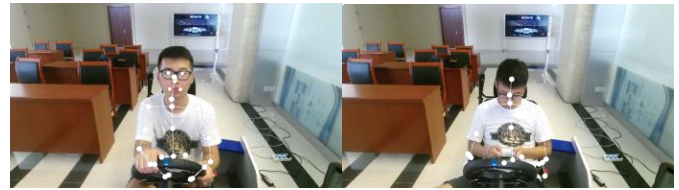

Fig. 8. DMS for driver workload and distraction detection. The left image shows the driver is driving the vehicle remotely, while the left image indicates the driver is performing secondary task (texting).

Furthermore, the parallel driving system is equipped with a remote-control platform. In this way, a human operator can take over the vehicle control authority when an emergency occurs and a take-over request is sent to the operator. A driver monitoring system (DMS) is developed for the remote driver, which is designed to prevent the driver from distracting and overload. A low-cost range camera named the Kinect is mounted in front of the human driver. It can provide RGB image and depth image. Meanwhile, it enables the detection of driver head rotation, facial status such as the eye closure and mouth open, and upper body joints [29].

As shown in Fig. 8, the driver workload is detected according to the mouth and eye states. If a long period of eye closure and frequent yawning was found, then a warning voice will be sent to the driver. Meanwhile, if the driver is detected as looking away or performing secondary tasks, the DMS will also warn the driver to concentrate on the current driving task until the real vehicle requests a takeover or the driving task is finished.

V. Concluding Remarks

This paper proposed the development of digital quadruplets for CPSS-based parallel driving. The four components of the proposed digital quadruplets are called descriptive, predictive, prescriptive and real vehicles. The goal of the descriptive vehicles is modeling and describing the real vehicles and recording percepts from the real environment. The predictive vehicles aim to predict the future dynamic driving environment and generate strategies in aspects of decision-making, planning and control, of the automated vehicles. The prescriptive vehicles decide how to interact with the real vehicles. Furthermore, the experimental validation of the digital quadruplets in parallel driving was introduced and the corresponding results were presented to demonstrate its effectiveness. The presented digital quadruplets in parallel driving will be expected to enable the future connected automated vehicles to drive more safely, efficiently and cooperatively. The framework of digital quadruplets for CPSS-based parallel driving still requires significant future research and development efforts, along with their real-world applications, which are being devoted by the research team.

Acknowledgement

The work is supported by National Natural Science Foundation of China (61533019, 91720000), Beijing Municipal Science & Technology Commission (Z181100008918007) and the Intel Collaborative Research Institute for Intelligent and Automated Connected Vehicles ("ICRI-IACV").

References

[1] F.-Y. Wang, "Parallel system methods for management and control of complex systems," *Control Decision*, vol. 19, no. 5, pp. 485–489, 2004.
[2] F.-Y. Wang, "Agent-based control for networked traffic management systems," *IEEE Intell. Syst.*, vol. 20, no. 5, pp. 92–96, 2005.
[3] F.-Y. Wang, "Artificial societies, computational experiments, and parallel systems: An investigation on computational theory of complex social economic systems," *Complex Syst. Complexity Sci.*, vol. 1, no. 4, pp. 25-35, 2004.
[4] T. Liu, B. Tian, Y. Ai, L. Li, D. Cao, and F.-Y. Wang, "Parallel reinforcement learning: a framework and case study," *IEEE/CAA Journal of Automatica Sinica*, vol. 5, no. 4, pp. 827-835, 2018.
[5] F.-Y. Wang, "Parallel control and management for intelligent transportation systems: Concepts, architectures, and applications," *IEEE Trans. on Intelligent Transportation Systems,* vol. 11, no.3, pp. 630-638, 2010.
[6] N. Zhang, F.-Y. Wang, F. Zhu, D. Zhao, and S. Tang, "DynaCAS: Computational experiments and decision support for ITS," *IEEE Intell. Syst.*, vol. 23, no. 6, pp. 19–23, 2008.
[7] F.-Y. Wang, "The emergence of intelligent enterprises: from CPS to CPSS," *IEEE Intelligent Systems*, vol. 25, no. 4, pp. 85-88, 2010.
[8] T. Liu, H. Yu, T. Bin, Y. Ai, and L. Chen, "Intelligent command and control systems for intelligent vehicle: primary methods and systemic construction," *Journal of Command and Control*, 2018.
[9] H. Wang, Y. Huang, A. Khajepour, "Cyber-physical control for energy management of off-road vehicles with hybrid energy storage systems," *IEEE Trans. on Mechantroincs*, vol. 23, no. 8, pp. 2609-2618, Dec. 2018.
[10] F.-Y. Wang, "Artificial intelligence and intelligent transportation: driving into the 3rd axial age with ITS," *IEEE Trans. Intell. Transp. Syst. Mag.*, vol. 9, no. 4, pp.6-9, 2017.
[11] F.-Y. Wang, N.-N. Zheng, et al., "Parallel driving in CPSS: a unified approach for transport automation and vehicle intelligence," *IEEE/CAA Journal of Automatica Sinica*, vol. 4, no. 4, pp. 577-587, Sep. 2017.
[12] F.-Y. Wang, J. Zhang, et al., "PDP: parallel dynamic programming," *IEEE/CAA J. Autom. Sinica.*, vol. 4, no. 1, pp.1-5, 2017.
[13] T. Liu, Y. Zou, et al., "Reinforcement learning of adaptive energy management with transition probability for a hybrid electric tracked vehicle," *IEEE Trans. Ind. Electron.*, vol.62, no.12, pp.7837-7846, Dec. 2015.
[14] T. Liu, X. Tang, H. Wang, H. Yu, and X Hu, "Adaptive Hierarchical Energy Management Design for a Plug-in Hybrid Electric Vehicle," *IEEE Trans. on Veh.Techn.*, vol. 68, no. 12, pp. 11513-11522, 2019.
[15] T. Liu, X. Hu, et al., "A heuristic planning reinforcement learning-based energy management for power-split plug-in hybrid electric vehicles," *IEEE Trans. on Ind. Inf.,* vol. 15, no. 12, pp. 6436-6445, 2019.
[16] F.-Y. Wang, "Computational experiments for behavior analysis and decision evaluation of complex systems," *Journal of system simulation*, vol. 16, no. 5, pp.893-897, 2004.
[17] T. Liu, H. Yu, H. Guo, Y. Qin, and Y. Zou, "Online energy management for multimode plug-in hybrid electric vehicles," *IEEE Trans. on Ind. Inf.,* vol. 15, no. 7, pp. 4352-4361, July 2019.
[18] U. Ritzinger, J. Puchinger, R. Hartl, "A survey on dynamic and stochastic vehicle routing problems," *Int. J. Prod. Res.*, vol. 54, pp. 215–31, 2016.
[19] N. Agatz, A. Erera, et al., "Dynamic ride-sharing: a simulation study in metro Atlanta," *Transp. Res. B*, vol. 45, pp. 1450–64, 2011.
[20] L. Li, Y. Lin, et al., "Parallel learning: a perspective and a framework," *IEEE/CAA J. Autom. Sinica.*, vol. 4, no. 3, pp. 389-395, July. 2017.
[21] J. Carrasquilla, R. Melko, "Machine learning phases of matter," *Nature physics*, vol. 13, pp. 431-434, 2017.
[22] T. Menies, "The Unreasonable effectiveness of software analytics," *IEEE software*, vol. 35, no. 2, pp. 96-98, Mar. 2018.
[23] P. Abbeel, and A. Ng, "Apprenticeship learning via inverse reinforcement learning," in *Proc. of the twenty-first international conference on Machine learning*, p. 1, July 2004.
[24] T. Liu, and X. Hu, "A bi-level control for energy efficiency improvement of a hybrid tracked vehicle," *IEEE Trans. Ind. Informat.*, vol. 14, no. 4, pp. 1616-1625, 2018.
[25] T. Liu, B. Wang, and C. Yang, "Online Markov Chain-based energy management for a hybrid tracked vehicle with speedy Q-learning," *Energy*, vol. 160, pp. 544-555, 2018.
[26] Y. Xing, *et al.* End-to-End Driving Activities and Secondary Tasks Recognition Using Deep Convolutional Neural Network and Transfer Learning. *IEEE Intelligent Vehicles Symposium*, pp. 1626-1631, 2018.
[27] Shariff, A., Bonnefon, J. F., & Rahwan, I. (2017). Psychological roadblocks to the adoption of self-driving vehicles. *Nature Human Behavior*, vo. 1, no. 10, 694.
[28] T. Liu, H.Wang, B. Tian, Y. Ai and L.Chen, "Parallel Distance: A New Paradigm of Measurement for Parallel Driving," *IEEE/CAA J. Autom. Sinica.*, vol. 1, no. 99, pp. 1-1, 2019.
[29] T. Liu, B. Tian, Y. Ai, F.-Y., Wang, "Parallel Reinforcement Learning-Based Energy Efficiency Improvement for a Cyber-Physical System," *IEEE/CAA J. Autom. Sinica.*, vol. 7, no. 2, pp. 617-626, 2020.